\documentclass[runningheads]{llncs}

\usepackage{microtype}
\usepackage{graphicx}
\usepackage{subfigure}
\usepackage{float}
\usepackage{booktabs} % for professional tables
\usepackage{physics}
\usepackage{blochsphere} 
\usepackage{amssymb}
\usepackage{hyperref}
\PassOptionsToPackage{hyphens}{url}
\usepackage{microtype}
\usepackage{tikz}
\usetikzlibrary{quantikz}

\usetikzlibrary{automata,positioning}

\PassOptionsToPackage{hyphens}{url}    
\usepackage{tikz,xcolor,hyperref}
\usepackage{microtype}
\usetikzlibrary{shapes,arrows,positioning}

\definecolor{lime}{HTML}{A6CE39}
\definecolor{prpa}{HTML}{648fff}
\definecolor{prpb}{HTML}{785ef0}
\definecolor{prpc}{HTML}{dc267f}
\definecolor{prpd}{HTML}{fe6100}
\definecolor{prpe}{HTML}{ffb000}

\DeclareRobustCommand{\orcidicon}{
	\begin{tikzpicture}
	\draw[lime, fill=lime] (0,0) 
	circle [radius=0.16] 
	node[white] {{\fontfamily{qag}\selectfont \tiny ID}};
	\draw[white, fill=white] (-0.0625,0.095) 
	circle [radius=0.007];
	\end{tikzpicture}
	\hspace{-2mm}
}
\foreach \x in {A, ..., Z}{%
	\expandafter\xdef\csname orcid\x\endcsname{\noexpand\href{https://orcid.org/\csname orcidauthor\x\endcsname}{\noexpand\orcidicon}}
}

\newcommand{\be}{\begin{eqnarray}}
\newcommand{\ee}{\end{eqnarray}}

\begin{document}

%\twocolumn[
%\icmltitle{
\title{A Review on Machine Learning Algorithms for Dust Aerosol Detection using Satellite Data}
\titlerunning{A Review on ML Algorithms for Dust Aerosol Detection using Satellite Data}

\author{Nurul Rafi
\inst{1}
\and
Pablo Rivas
\inst{2}
\orcidC{} 
}

\institute{
School of Engineering and Computer Science \\
Department of Computer Science \\
Baylor University, Texas, USA\\
$^1$\email{Nurul\_Rafi1@Baylor.edu}
$^2$\email{Pablo\_Rivas@Baylor.edu}
}
\maketitle
\begin{abstract}
Dust storms are associated with certain respiratory illnesses across different areas in the world. Researchers have devoted time and resources to study the elements surrounding dust storm phenomena. This paper reviews the efforts of those who have investigated dust aerosols using sensors onboard of satellites using machine learning-based approaches. We have reviewed the most common issues revolving dust aerosol modeling using different datasets and different sensors from a historical perspective. Our findings suggest that multi-spectral approaches based on linear and non-linear combinations of spectral bands are some of the most successful for visualization and quantitative analysis; however, when researchers have leveraged machine learning, performance has been improved and new opportunities to solve unique problems arise. 
\end{abstract}

\section{Introduction}

Dust is the most common form of aerosol globally, affecting the water cycle, plants, public health and welfare, and climate \cite{namdari2018impacts}. It is generated at a micro-scale and can affect a wide area depending on wind flow and geomorphology \cite{boroughani2020application}. Dust aerosols are non-spherical airborne particles with depolarization and can be found in large numbers, particularly in areas like Africa's northwestern region. However, researchers discovered that dust aerosols could be found around different continents, regardless of their source \cite{rivasnert}. 

\begin{figure}[t!]
 \centering
 \includegraphics[width=\textwidth]{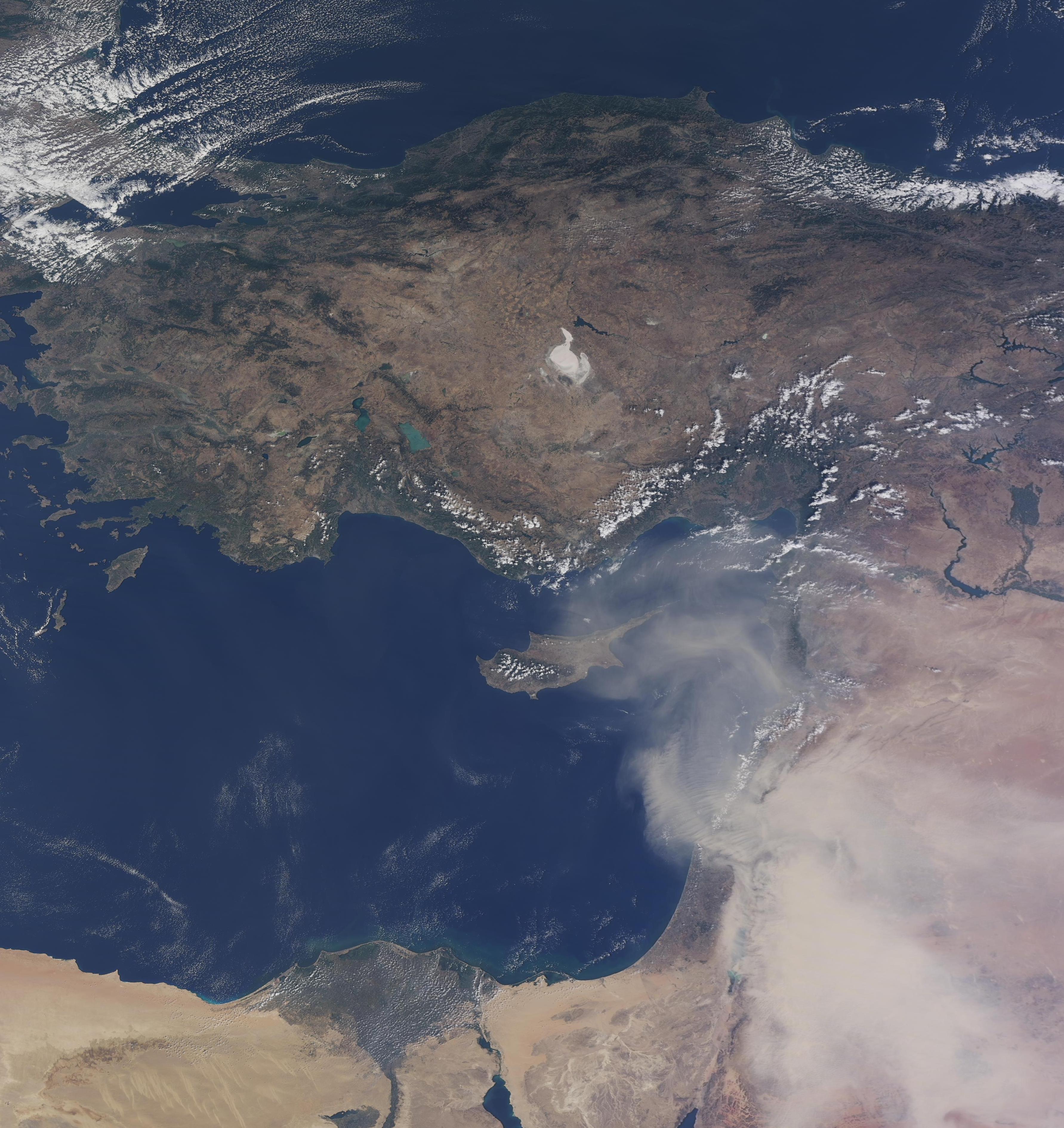}
 \caption{Dust event captured with the MODIS instrument over NASA's Terra Satellite. Source: West Africa.}
 \label{fig:dust}
\end{figure}

According to some reports, dust aerosol causes extreme air pollution, and a variety of respiratory diseases \cite{rivas2010probabilistic,rivera2006detection}. Dust storms, which contain toxic airborne particles such as organic contaminants, trace products, and cancer-causing bacteria, are deadly weather phenomena that mostly occur in deserts and bare land areas \cite{shi2020developing,ahn2008comparison,erel2006trans,kaiser2005mounting}. Dust storms directly impact the global environment by absorbing solar radiation and reducing visual acuity, resulting in dangerous traffic accidents \cite{bishop2002robotic,alizadeh2014global}. To assess the level of activity of dust storms, researchers evaluate changes in different criteria, including dust day's frequencies \cite{shao2003climatology,ekhtesasi2009investigation}, optical depth index of aerosols \cite{butt2017assessment}, and index of a dust storm \cite{o2014dust,ebrahimi2021evaluation}, among others \cite{ma2015transfer}. While some of these dust events are evidently visible, as shown in Figure \ref{fig:dust}, low concentrations of dust at different altitudes can present challenges that require leveraging machine learning methodologies for better results. This paper performs a succinct literature review of machine learning methodologies applied in dust aerosol modeling problems.

This paper is organized as follows: Section 2 introduces the concept of remote sensing for dust aerosols, including the different satellites and sensors commonly used. Section 3 describes data availability and accessibility-related issues. Section 4 presents approaches based on the physical properties of dust, paving the way to understand the feature space of machine learning methodologies, which are discussed in Section 5.  Section 6 provides additional information on methods related to dust modeling, and finally, conclusions are drawn in Section 7.

\section{Remote Sensing for Dust Aerosols}
The use of remote sensing (RS) to detect dust sources at both local and regional scales is a valuable tool \cite{baddock2011geomorphology}. Due to the high variability in spatial data, RS has become the standard method for determining the presence and movement of dust aerosols.
In addition to AERONET \cite{kim2007seasonal} and lidar systems \cite{ruckstuhl2009aerosol}, specific RS may be useful depending on the target's characteristics and location \cite{badarinath2010long}.
The majority of satellites are polar orbiting, with higher spectral resolution and a wider range, while geostationary satellites, with their high temporal resolution \cite{shi2020developing}, are used to monitor the formation and growth of dust storms.
Geostationary satellites have more channels and resolution for collecting dust temporal variations \cite{li2020review}.

\subsection{Aqua and Terra: MODIS}
Four indicators from Moderate Resolution Imaging Spectroradiometer(MODIS) satellite images from both Terra and Aqua satellites \cite{rivas2010probabilistic} have been used in the study of RS around the world to identify dust sources, including Brightness Temperature Differences (band 29–band 31) and (band 31–band 32), as well as the Normalized Difference Dust Index (NDDI) and D variable using Over-land \cite{miller2003consolidated,boroughani2020application}. To better understand dust storms, some researchers use MODIS Terra Level 1B radiances to capture images near real-time with 1km resolution. Some models, on the other hand, use the MODIS Aerosol Optical Thickness (AOT) product, which has a 10km resolution and takes several hours to process after obtaining images \cite{rivasdust}. The concentration of the AOD product is high, but the spatial resolution is poor \cite{shahrisvand2013comparison}. MODIS AOT also ignores fine dust and does not operate in heavy icy clouds \cite{zhao2010dust}. NASA's EOS (LANCE) offers data almost in real time \cite{rivasnert}, and is stacked in this website where anybody can search and collect data, with data from MODIS being called a granule every 5 minutes \cite{shi2019hybrid}. 

MODIS data is divided into three levels: Level-0, 1A, and 1B, with Level-1B containing corrected multi-spectral data \cite{rivas2010automatic}. The 16 bit thermal emitting bands in 1B level data should be processed to this $Wm^{-2}\mu m^{-1}sr^{-1}$ unit \cite{kopp2011new}. With this 3 level info, MODIS provides 36 spectral bands with a spatial resolution of 250m(1-2), 500m(3-7), 1km(8-36) bands \cite{shi2020developing} and a wavelength range of 0.045 to 14.385$\mu m$ and an instrument angle of 110 degree \cite{badarinath2010long}. Raw telemetry data can be found on Level 0 and raw spectral bands data can be found on Level 1A \cite{rivas2010automatic}. For RGB mapping, the bands $1, 4$ and $3$ are used \cite{rivas2013statistical}.

\subsection{CALIPSO}
Cloud–Aerosol Lidar and Infrared Pathfinder Satellite Observation (CALIPSO) has a dust dataset with a high frequency. Using its vertical profile \cite{li2020review} and a polar orbiting satellite with a 16-day recycle period \cite{someya2019dust} and a 5-kilometer range, it can provide more detailed dust information. Dust storms were collected using CALIPSO \cite{huang2007summer}. CALIPSO can detect vertical distribution in normal time \cite{badarinath2010long}. Though CALIPSO is better at detecting dust, it only collects data from a smaller area of the Earth's surface \cite{shi2019hybrid}. Aside from that, it has a high consuming power and a small overage field \cite{shi2020mineral}. 

CALIPSO uses a 98°-inclination orbit to test lidar signals in the 532 and 1064 $nm$ bands \cite{shi2019hybrid} and flies at an altitude of 705 km to provide vertical distribution of aerosols and cloud \cite{badarinath2010long}. Based on the probability density function, \cite{liu2014discrimination} used CALIPSO data to describe the dust aerosol. 

\subsection{VIRRS}

At night, the Visible Infrared Imaging Radiometer Suite loses its solar reflective bands and has 22 channels with wavelengths ranging from $0.41\mu m$ to $12.01\mu m$ \cite{lee2021machine}. 16 have moderate resolution imagery bands, 5 have high resolution imagery bands, and one is a panchromatic day/night band.  

\subsection{CALIOP}
The Cloud-Aerosol Lidar with Orthogonal Polarization is a lidar device in space that can measure depolarization and color ratio \cite{ma2015transfer}. At $532 nm$ and $1064 nm$, CALIOP offers backscatter profiles as well as two orthogonal (parallel and perpendicular) polarization components at $532 nm$ \cite{badarinath2010long}. With the aid of other radiation models, CALIOP data can be useful in achieving radiative and heating rates \cite{huang2009taklimakan}. Versions 3.01 and 3.02 of the CALIOP Level 2-5 km Aerosol/Cloud layer were used in this paper\cite{someya2019dust}. 

\subsection{MERIS}
Dust samples from sand regions were collected using Medium Resolution Imaging Spectrometer (MERIS) \cite{wei2008detection,chacon2011dust}. Ocean color is also collected by MERIS.

\subsection{AVHRR}
The AVHRR bands 4 and 5 are the same as MODIS 31 and 32 \cite{rivera2006detection}. The NOAA-AVHRR satellites' GAC (Global Area Coverage) data set is an excellent resource for researching in surface conditions caused in part by climatic variation. AVHRR  has  5 spectral bands varying from $0.58 - 12\mu m$ \cite{fjeldsaaa1997biodiversity}. 

\subsection{OMI}

Researchers used the OMAERUV product from the Ozone Monitoring Instrument (OMI) sensor to verify the accuracy, as they used the AQUA satellite where OMI is appropriate \cite{souri2015dust}. It has a high spatial resolution and is used to detect both sources and plumes \cite{badarinath2010long}.

\section{Data Sources and Computer Processing}

\subsection{Data Collection}
Dusty days are described as days with a visibility of less than 2000 meters per hour during the day, as measured by Terra-MODIS images \cite{boroughani2020application,rashki2012dust,rashki2013dryness,vickery2013dust}. Terra-MODIS can only generate 23 images per day due to its daily visit to a predetermined location \cite{boroughani2020application}. The MODIS sensor from Terra and Aqua is used for a variety of purposes, including (i) wide angle and high temporal resolution of images (ii) high spatial resolution with various band ranges iii) a wide variety of spectral bands and ranges  \cite{engelstaedter2003controls,mahowald2003ephemeral,walker2009development,hahnenberger2014geomorphic}.

Authors used webscraping to gather data from NASA at various resolutions and times \cite{rivasnert}. Researchers used smaller feature vectors with KLT due to high computational complexity and hardware dependence. In order to take at least 30 times training samples of the feature bands, they chose 240 feature vectors based on these two groups. \cite{mather1997geological,rivas2013statistical}. They implemented Balanced Error Rate (BER) as a method of mitigating error to eliminate false positive counts. Bias correction can be used to eliminate inconsistencies in results \cite{jin2019machine}.

\subsection{Data Access} 
Table \ref{tbl:data} contains relevant information for accessing data from NASA information systems.

\begin{table}[h]
    \centering
    \caption{Data Sources by NASA}
    \begin{tabular}{|p{3.5cm}|p{5cm}|p{3.5cm}|} \hline
        Source & URL & Description \\ \hline
        LAADS DAAC Level-1 and Atmosphere Archive \& Distribution System Distributed Active Archive Center & \url{https://ladsweb.modaps.eosdis.nasa.gov} & MODIS, VIIRS, MERIS, SLSTR, and OLCI are all described in detail. \\ \hline
        Terra and Aqua & \url{https://ladsweb.modaps.eosdis.nasa.gov/missions-and-measurements/modis/} & Terra and Aqua are described in detail, with all technical requirements and channels containing all band information.\\\hline 
        VIIRS - Visible Infrared Imaging Radiometer Suite & \url{https://ladsweb.modaps.eosdis.nasa.gov/missions-and-measurements/viirs/} & All technical requirements and channels for VIIRS are listed, as well as all band details.\\ \hline
        MODIS Level 0-1 & \url{https://ladsweb.modaps.eosdis.nasa.gov/missions-and-measurements/science-domain/modis-L0L1/} & Both levels of data are accessible in this link's data repository. A total of 8 Terra and Aqua products, as well as MODIS Level 1B calibrated data with spatial resolutions of 250m, 500m, and 1km, are available here.\\ \hline
    \end{tabular}
    \label{tbl:data}
\end{table}

\begin{figure}[hbt!]
 \centering
 \includegraphics[width=\textwidth]{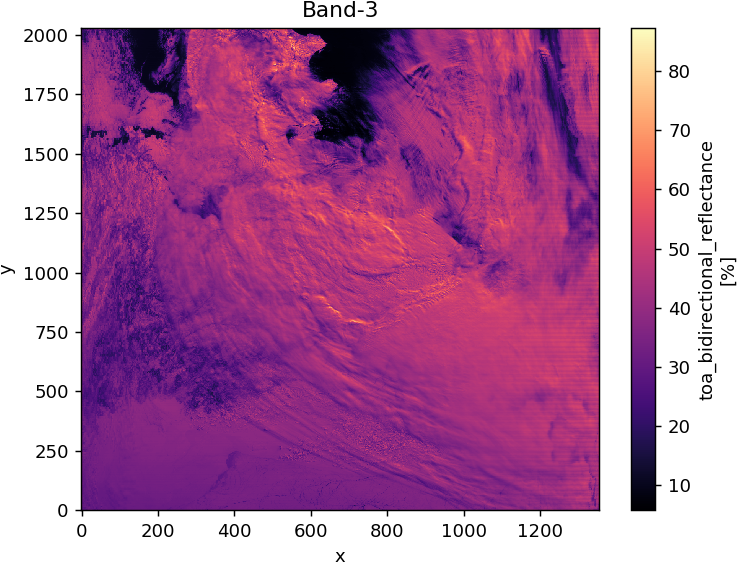}
 \caption{Band-3 in magma color map}
 \label{fig:band3}
\end{figure}

MODIS contains hdf5 files where all the information of the bands are stacked in this format. We have used the file {\tt  MOD021KM.A2021092.0020.006.2021092134055.hdf} for data processing and for band information. This is 1km spectral resolution file as the prefix of the file name defines it. In Figure \ref{fig:band3}, only band 3 is shown in a magma color map using the Python language. For processing hdf5 file and band information Satpy Python package was used.

\section{Physical approaches}
\subsection{MODIS Index}
They introduced Brightness Temperature Difference for dust detection in this paper \cite{ackerman1989using}. Positive BTD at a high level indicates dense dust, while positive BTD at a low level may also indicate dense dust \cite{souri2015dust}. $B T D_{3-11}$ is sensitive to dust loading, while $B T D_ {8-11}$ and $B T D_{11-12}$ will distinguish the dust when used together \cite{li2020review}.

\begin{equation}
\begin{array}{l}
B T D_{11-12}=B T_{\sim 11.2}-B T_{\sim 12.4} \\
B T D_{3-11}=B T_{\sim 3.9}-B T_{\sim 11.2} \\
B T D_{8-11}=B T_{\sim 8.6}-B T_{\sim 11.2}
\end{array}
\end{equation}

In the Saharan zone, researchers developed TDI to detect dust \cite{huang2007summer}. They observed dust storms using NDDI indexing from MODIS data, and four bands were used. Dust was separated from water, clouds, and snow using various thresholds \cite{xie2009detection}.
\begin{equation}
N D D I=\frac{B 7-B 3}{B 7+B 3}
\end{equation}

NDDI is commonly used in conjunction with BTD because it cannot detect dust on its own \cite{li2020review}. For measuring dust events, various types of indices are used, and multiple indices can be combined to resolve drawbacks and perform better \cite{shi2020developing}. The BADI and BTR methods eliminate some of the drawbacks of the BTD method \cite{shi2020developing}. Since it only uses the thermal wavelength, TDI can be used at night as well as during the day while it is over the ocean \cite{li2020review}

\begin{equation}
\mathrm{TDI}=\mathrm{C}_{0}+\mathrm{C}_{1} \times \mathrm{B} \mathrm{T}_{3.7}+\mathrm{C}_{2} \times \mathrm{B} \mathrm{T}_{9.7}+\mathrm{C}_{3} \times \mathrm{B} \mathrm{T}_{11}+\mathrm{C}_{4} \times \mathrm{B} \mathrm{T}_{12}
\end{equation}

BADI is more effective at catching the size and density of dust \cite{li2020review}.
\begin{align}
\mathrm{BADI}=&\frac{2}{\pi} \times \arctan \left(\frac{\mathrm{BDI}}{B D I_{0.95}}\right), \\
\text { where } \mathrm{BDI}=&\left(B T D_{3.9-11.2}\right)^{2} \times B T D_{12.4-11.2}.
\end{align}

\subsection{Generic}
Low level visual grammar scheme describes high level spatial feature \cite{aksoy2005learning}. The majority of them avoided feature engineering \cite{shi2019hybrid}.

\section{Machine Learning Algorithms}

\subsection{Support Vector-Based}
SVM needs a lot of memory and can struggle with large datasets \cite{shi2019hybrid}. SVM maximizes the decision boundary of two groups and embraces non-linear classification, but it takes more time to process, which is inconvenient for large datasets \cite{shi2020developing}. SVM attempts to find the optimal hyperplane \cite{nabavi2018prediction} by converting non linear problems to linearly separable problems using kernel functions \cite{zhang2004wavelet}. On their dataset, the authors claimed a precision of up to 98 percent \cite{rivas2015near}. They used four bands, B20, B29, B31, and B32, and tried to reduce the support vectors to improve the dust event likelihood.

They used TrAdaBoost for SVM transfer learning above 3km altitude with high classification accuracy \cite{ma2015transfer}. SVM for Regression was used to detect dust aerosols in near real time, with the researchers attempting to reduce the time complexity by minimizing the support vectors \cite{rivas2015near}. Among the other approaches, Large Scale Linear Programming-SVR performed better in classifier efficiency. The rank of the classifiers is indicated in parenthesis.

Researchers used updated SVM with MODIS L1 data because threshold-based SVM has some disadvantages in terms of complexity and certainty. They discovered that previous methods had difficulty correctly distinguishing dust on bright surfaces. They used the Radial Basis Function to find the best classification observation (RBF) \cite{shi2020developing}. Since it is dependent on the input data, bias may affect the model's output \cite{li2020review}.

In a classification problem, SVM outperforms ANN because it always finds the global minimum. When the regularization parameter $C = 100$ and the kernel $gamma=0.008$ are both chosen for classification, RBF is the most efficient kernel. Auxiliary data, such as cloud masking, is not required for SVM.
SVM worked well in both land and oceans with low dust density, according to their findings. However, they mistook the majority of cloudy pixels for dust. The precision was calculated using the kappa coefficient.
They also claimed that SVM is a sufficient tool for dust detection in multispectral images. SVM had an accuracy of 84 percent with a Kappa coefficient of 0.8091, while MLP had an accuracy of 81 percent with a Kappa coefficient of 0.771 \cite{shahrisvand2013comparison}. C and $gamma$ are commonly used values of 1 and 0.07, respectively, for dust aerosol detection with SVM and CALIPSO data, while C(0.25) and $gamma$(0.0078) are used for dust aerosol detection with SVM and CALIPSO data \cite{shi2020developing,ma2012evaluating}. 

SVM was used to identify the best band combinations, which were B7-B3, B20-B31, and B31-B32 \cite{shi2020developing}. Although SVM has the advantage of using small samples, 5D PDF performed better after the first day of observation \cite{ma2015transfer}. Also, for high dimension or huge number of data, training SVM is difficult \cite{jiao2021next}.

\subsection{Neural Networks}

\subsubsection{Feed-Forward or Dense}
The input, secret, and output layers of an artificial neural network are typically represented by feature vectors. Researches proposed for detecting clouds present in the B3 and B7 bands in order to minimize the high complexity of dust image processing. For the gradient, they used the Scaled Conjugate Gradient Algorithm because it is more efficient. The classifier identified three classes: vegetation, soil, and dust \cite{chacon2011dust}. In their study, they used the R packages Neuralnet and SparkR to perform ANN analysis \cite{shi2020mineral}. The neural network was fed the reflection values of the bands as data \cite{shahrisvand2013comparison}. The authors recommend two hidden layers, but the most appropriate number of layers and parameters can be determined through repetition and comparison of the best accuracy \cite{mather2016classification}.

Feed Forward Neural Networks (FFNN) functioned similarly to a traditional neural network, with input and output layers as well as hidden layers. Back propagation and the "Levenberg–Marquardt" algorithm were used by the researchers to update weights and prejudice \cite{rivas2013statistical}. The error was calculated using the mean squared error (MSE) where between the layers, there is no cycle or loop. 3 secret layers, each with a batch size of 256 and a total of 2000 epochs \cite{lee2021machine}. Among the models that used CALIOP products, FFNN had an accuracy of 84.9 percent \cite{lee2021machine}. In FFNN, they used sigma as an activation function \cite{rivasdust}. 

The researchers used a two-layer Feed-Forward neural network with 15 secret neurons and three neuron outputs \cite{chacon2011dust}. They simplified the process of choosing statistical datasets and eliminating false positives.

\subsubsection{Convolutional}
Convolutional Neural Network (CNN) is more suitable for image processing. CNN applies filters to reduce the dimensionality of images without eliminating essential features \cite{lee2021machine}. SVM has some limitations due to the high dimensionality or huge amount of samples, while ANN has issues with false positives. For dust detection, a developed Naive Bayesian CNN classification technique is being developed. Because $PM_{2.5}$ concentration is linked to dust, the CNN method is utilized to forecast $PM_{2.5}$ value \cite{jiao2021next}. Although the Softmax classifier is utilized for final layer classification in CNN, it is insufficient for splitting the feature space due to high noise in remote sensing data or huge differences between features \cite{song2019survey}. 

In remote sensing images, CNN outperformed traditional methods. Atrous convolution is used in conjunction with separable convolution because it captures a broader context with its larger view of filters, which may perform better in scene categorization in remote sensing \cite{chen2020convolution}. In order to improve accuracy, they merged CNN and advanced MLP in the classification step \cite{shawky2020remote}. Based on ResNet101, the fast deep perception network (FDPResnet) includes deep CNN and Broad Learning System (BLS) for extracting both deep and shallow aspects of remote sensing images. The proposed method performed well in classification in a shorter time using the NWPU-RESISC45 remote sensing dataset \cite{dong2020fast}. As attention mechanism identifies the most active region in a remote sensing image for better classification \cite{wang2018scene} first introduced Attention Recurrent Convolutional Network (ARCNnet).

On six popular remote sensing datasets, they applied a new EfficientNet CNN model based on attention mechanism to the last feature map.
EfficientNet CNN outperformed earlier approaches, and they employed the Swish activation function, which is smoother than ReLU and LeakyReLU \cite{alhichri2021classification}. \cite{li2019deep} applied same model with different training rounds (SMDTR) with CNN which performed better than CNN, CapsNet, SMDTR\_CapsNet. CNN was utilized in conjunction with CapsNet, with CNN converting input images into feature maps and CapsNet performing final classification.
This integrated strategy outperformed other current approaches on three popular datasets \cite{zhang2019remote}. 

\subsubsection{Probabilistic}

The Probabilistic Neural Network (PNN) is a neuro-statistical hybrid model \cite{rivas2013statistical}. PNN is a semi-supervised network since it does not involve learning but operates under supervision \cite{specht1988probabilistic,chettri1993probabilistic}. It is employed in the classification of dust events. They used a minimized version of feature bands based on the assumption that training samples should be at least three times the size of feature bands due to the large number of feature vectors \cite{ramakrishnan2007image}. For dimension reduction, Principal Component Analysis (PCA) was used. They also used pixel mapping to address dust likelihood in areas farther away from the dust storm source \cite{rivas2010traditional}. It has four layers: input, pattern, summation, and output, with the summation layer centered on individual pattern layers in different groups \cite{rivas2010automatic}.

The layer performance of a pattern is determined by \cite{rivas2013statistical}
\begin{equation}
\varphi_{j k}(\boldsymbol{F})=\frac{1}{(2 \pi)^{\frac{d}{2}} \sigma^{d}} \mathrm{e}^{-\frac{1}{2 \sigma^{2}}\left(\boldsymbol{F}-v_{j k}^{F}\right)^{T}\left(\boldsymbol{F}-v_{j k}^{F}\right)},
\end{equation}
where $\sigma$ is the absolute difference between the smallest normal variances \cite{rivas2013statistical}. 

Researchers discovered that the Gaussian distribution better describes the frequency of dust storms using MODIS 1B data. Their probabilistic model was also capable of segmentation using the threshold value. They used the formula below to convert units back to the original unit. \cite{rivas2010probabilistic}. 
\begin{equation}
\hat{X}=\alpha(\beta-\gamma)
\end{equation}
Where $\alpha$ and $\beta$ represent radiance scales and offsets, respectively, and $gamma$ represents scaled data intensities.

PNN outperformed the ML classifier in this study \cite{rivas2010automatic}. Multi spectral images can be categorized using a probabilistic density function \cite{rivasdust}. Their probability density function was dubbed the "data likelihood" function by them \cite{rivas2010automatic}.

\subsection{Ensamble Method}
\subsubsection{Random forest}

For the first time, researchers used Random Forest (RF) for dust classification, except the cloud mask, which was a source of misclassification in previous approaches \cite{souri2015dust}. It outperforms SVM, AdaBoost, and ANN as an ensemble system. Also with the thin dusts, RF outperformed and in both water and ground \cite{xie2009detection} \cite{zhao2010dust}. RF can accommodate a large number of inputs without the use of variables \cite{shi2019hybrid}. However, their biggest flaw is that they used manual data to train RF, which should have been automated. 

It functions best with high-dimensional data because it works with subsets and also determines the value of features, reducing the difficulty of identifying predictor variables. In this study, 100 max depths and 500 estimators were used \cite{lee2021machine}. Susceptibility mapping was developed from different dust events for classification, and AOC 90.8 RF was performed \cite{boroughani2020application}.  For dust detection, a random forest model utilizing the Advanced Baseline Imager (ABI) achieved an AUC of 0.97 \cite{berndtmachine}. 

\subsection{Clustering}
\subsubsection{K nearest neighbours}
The main benefit is that data does not have to be linearly separable.
Hyperparameters were calculated using 10 nearest neighbors using the Euclidean distance calculation \cite{lee2021machine}.
\cite{proietti2015dust} used a k-NN classifier to identify dust in museum locations.
The size and shape of the particles are determined using a classification technique where the accuracy is more than 90\%. 
% \subsubsection{Kmeans}

\subsection{Maximum Likelihood-Based}

Maximum Likelihood Classifier is used for the classification of atmospheric components. It is based on a Bayesian classifier with probabilistic arguments, and two random variables, dust and backgrounds, were chosen \cite{rivas2013statistical}. They also counted the two occurrences as having an equal chance of occurring, and then determined the function vector $F$ that is closest to the classes \cite{rivas2013statistical}.

\begin{equation}
\psi_{c_{i}}(x)=\left(x-\boldsymbol{\mu}_{i}\right)^{T} \boldsymbol{\Sigma}_{i}^{-1}\left(x-\boldsymbol{\mu}_{i}\right)-\operatorname{det}\left(\boldsymbol{\Sigma}_{i}\right)
\end{equation}
Where $\mu_i$ is the mean feature vector and $\Sigma_i$ is the covariance matrix, and $\mu$ and $x$ are derived using ML estimators from training data.
The determinant function is "det," and the discriminant function is $\psi_{c_ i}$, which is quadratic and calculates the distance between $\mu$ and $x$ weighted by $\Sigma_i$. 

Finally, the feature vector will be classified into the two classes that are the most similar. The researchers concluded that the likelihood of dust and background groups was roughly equal. To find the discriminant function and simplify the problem, the log-likelihood of the Gaussian distribution was used. \cite{rivas2013statistical}. For satellite image classification, L2 regularization is used with logistic regression \cite{lee2021machine}.

\subsection{Comparison}

Machine learning methods are ideal for fusing because they allow for the use of a variety of inputs \cite{li2020review}. There is a comparison in the Table  \ref{tab:my_label} where all of the machine learning models in dust detection are listed from different researches along with precision, accuracy, AUC, processing time (PT), and RMSE. 

The image size in this experiment was $2030\times1053$. Around 97.5 percent of the 75 million feature vectors were from the context class, with 0.005 percent used for training and the rest for testing. A total of 24 training samples were used per class, with a processing time of 2.5 seconds \cite{rivas2010traditional,rivas2010automatic,rivas2010probabilistic}.

\begin{table}[t]
    \centering
    \caption{Performance Comparison}
    \begin{tabular}{lllllll}\hline
Model & Ref & Precision & Accuracy & AUC & PT & RMSE \\ \hline 
ML & \cite{rivas2010traditional,rivas2010automatic,rivas2010probabilistic} & 0.5255 & 0.6779 & 0.4884 & 0.0141 \\
PNN & \cite{rivas2010traditional,rivas2010automatic,rivas2010probabilistic} & 0.8080 & 0.8816 & 0.7035 & 0.2393 \\
FFNN & \cite{rivas2010traditional,rivas2010automatic,rivas2010probabilistic} & 0.7664 & 0.8412 & 0.6293 & 0.0459 \\
LP-SVR & \cite{rivas2010traditional,rivas2010automatic,rivas2010probabilistic} & 0.7907 & 0.8678 & 0.7117 & 0.0809 \\
LS LP-SVR & \cite{rivas2010traditional,rivas2010automatic,rivas2010probabilistic} & 0.8295 & 0.9104 & 0.7349 & 0.0974 \\
PD & \cite{rivasdust} & 0.3938 & 0.4964 & 0.4993 & 0.0141 \\
FFNN & \cite{rivasdust} & 0.4554 & 0.5426 & 0.7402 & 0.0472 \\
RF & \cite{rahmati2020identifying,nabavi2018prediction,shi2019hybrid,shi2020developing} &  & 0.798  & 0.894 & & 0.16 \\
SVM & \cite{rahmati2020identifying,nabavi2018prediction,shi2019hybrid,shi2020developing} &  & 0.658  & 0.875 & & 0.273\\
MARS & \cite{rahmati2020identifying,nabavi2018prediction} &  &  & 0.81 & & 0.315\\

LR & \cite{shi2019hybrid,shi2020developing} &  & 0.839  & 0.864 \\
Ensemble & \cite{shi2019hybrid,shi2020developing} &  & 0.756  & 0.68 \\ \hline
\end{tabular}
    \label{tab:my_label}
\end{table}

The optimal SVR parameters for 38 dust events with 97 million feature vectors were $\sigma$ = 0.125, $\eta$ = 0.5, and $\epsilon$ = 0.1 \cite{rivas2015near}, where they used 42 million four-dimensional samples with 92 percent precision in this paper \cite{rivasnert}. Neural PNN outperforms FFNN, and LP-SVR is the highest, with statistical LP-SVR outperforming ML classifier, which has a false positive rate of more than 50\% \cite{rivas2013statistical}. They discovered that the output of PNN and LP-SVR is not substantially different when they calculated the crucial difference \cite{rivas2013statistical}. TSS assesses all commission and omission mistakes. RF had the highest TSS value, while MARS had the lowest \cite{rahmati2020identifying}.

There are the advantages and disadvantages of using a physical approach over a machine learning approach (55 vs 93 for same day prediction, and 42 vs 80 for different day prediction). The processing time was reduced from 30 hours to 30 minutes thanks to the High Performance Computing Facility (HPCF) \cite{shi2020mineral}.

The models' success in predicting the winter dust storm index in Iran's arid regions was addressed \cite{ebrahimi2021evaluation}. They analyzed Dust Storm Index using MARS, LASSO, k-NN, GP, SVR, Cubist, RF, XGB, DNN, and GRA in the arid area of Iran, and found GRA to be the most accurate where DSI indicates wind erosion.

They measured the efficiency of different machine learning algorithms using AUC and true-skill statistics (TSS). RF was the most successful in deciding that the most critical factors in dust generation were wind speed and land cover \cite{rahmati2020identifying}.

\section{Other Methods}
\emph{Band-math} is a statistical illustration-based method for detecting dust events \cite{rivas2010traditional}.
Different types of sky can result in different color ratios, which were used to detect dust. Cloud reflectance and emission properties can also be used to observe dust aerosols \cite{shi2020mineral}.
They used a neuro-fuzzy inference method that was adaptive (ANFIS). Multivariate and adaptive regression splines were also used (MARS) \cite{rahmati2020identifying}.

% \section{Open Problems} TODO: this section is completely unrelated to any evident open problems
Researchers used neural networks and SVM to correct bias between MODIS AOD and AERONET AOD. Since SVM has the property of operating in higher dimensional space with kernel mapping, it outperformed neural networks in bias correction of MODIS Terra and Aqua \cite{lary2009machine}.

\section{Conclusions}
This paper has revised the most common methodologies that are applied to model dust aerosols for different purposes. Dust aerosols can be modeled for detection, probability estimation, discrimination among other aerosols, and even segmentation of dust storms. The general instrument is satellite imagery, which can take different forms depending on the hardware and sensors used to capture atmospheric phenomena. Our study found that the most common sensing type is multispectral, providing a rich spectral signature for different atmospheric analysis tasks. 

The most successful algorithms for dust aerosol modeling in earlier work have been based on a straight-forward linear combination of spectral bands. However, with recent advances in machine learning, a lot of work has been successful despite the common explainability critiques. Efficient deep learning models are very good at modeling highly complex phenomena, and it seems like this area requires more exploration. Here are some areas that we identified: 
i) attention-based recurrent models for exploring dust aerosols over time;
ii) convolutional approaches for hyperspectral cubes; and 
iii) hybrid attention-based deep learning and semi-supervised approaches.

Future work will examine some of these possible areas of experimentation and research the potential advantages of leveraging novel deep learning models.

%\section*{Acknowledgements}
%The authors thank the **omitted for blind review** for their support under grant **omitted for %blind review**. This research was also funded, in part, by **omitted for blind review**.

\bibliographystyle{splncs04}
\bibliography{example_paper}

\end{document}